\documentclass[lettersize,journal]{IEEEtran}
\usepackage{amsmath,amsfonts}
\usepackage{algorithmic}
\usepackage{algorithm}
\usepackage{array}
\usepackage[caption=false,font=normalsize,labelfont=sf,textfont=sf]{subfig}
\usepackage{textcomp}
\usepackage{stfloats}
\usepackage{url}
\usepackage{verbatim}
\usepackage{graphicx}
\usepackage{cite}
\usepackage{xcolor}
\usepackage{multirow}
\usepackage{amsfonts}
\usepackage{pifont}
\usepackage{amssymb}
\usepackage{graphicx}
\usepackage{float}

\newcommand{\etal}{\textit{et al. }}
\newcommand{\cmark}{\ding{51}}%
\newcommand{\xmark}{\ding{55}}%

\begin{document}

\title{Online Video Super-Resolution with Convolutional Kernel Bypass Graft}

\author{Jun Xiao, Xinyang Jiang, Ningxin Zheng, Huan Yang, Yifan Yang, Yuqing Yang, Dongsheng Li, Kin-Man Lam
\thanks{Jun Xiao and Kin-Man Lam are with the Department of Electronic and Information Engineering, the Hong Kong Polytechnic University. E-mail: jun.xiao@connect.polyu.hk.}

\thanks{Xinyang Jiang, Ningxin Zheng, Huan Yang, Yifan Yang, Yuqing Yang, and Dongsheng Li are with Microsoft Research Asia. E-mail: xinyangj@hotmail.com}

\thanks{Most work of this paper were finished when Jun Xiao interned in Microsoft Research Asia.}






}

\markboth{Journal of xxx,~Vol.~14, No.~8, August~2022}%
{Shell \MakeLowercase{\textit{et al.}}: A Sample Article Using IEEEtran.cls for IEEE Journals}


\maketitle

\begin{abstract}
Deep learning-based models have achieved remarkable performance in video super-resolution (VSR) in recent years, but most of these models are less applicable to online video applications. These methods solely consider the distortion quality and ignore crucial requirements for online applications, e.g., low latency and low model complexity. In this paper, we focus on online video transmission, in which VSR algorithms are required to generate high-resolution video sequences frame by frame in real time. To address such challenges, we propose an extremely low-latency VSR algorithm based on a novel kernel knowledge transfer method, named convolutional kernel bypass graft (CKBG). 
First, we design a lightweight network structure that does not require future frames as inputs and saves extra time costs for caching these frames. 
Then, our proposed CKBG method enhances this lightweight base model by bypassing the original network with ``kernel grafts'', which are extra convolutional kernels containing the prior knowledge of external pretrained image SR models. In the testing phase, we further accelerate the grafted multi-branch network by converting it into a simple single-path structure. Experiment results show that our proposed method can process online video sequences up to 110 FPS, with very low model complexity and competitive SR performance. 
\end{abstract}


\section{Introduction}
\IEEEPARstart{V}{ideo} super-resolution (VSR) is a fundamental task in computer vision, which aims to generate high-resolution (HR) video sequences given the corresponding low-resolution (LR) counterparts. In general, VSR is a challenging problem because of its ill-posed nature, which means that an LR frame can be generated from infinitely possible HR frames. With the fast development of video applications in the last decade, VSR has shown substantial industrial values and thus attracted researchers' attention.

Online video applications (e.g., cloud gaming, live broadcasting, and online video conferences) have become increasingly popular, especially during COVID-19. Although existing deep learning-based models have achieved unprecedented success in VSR tasks, most of them are less applicable in online situations because they solely pursue performance improvement and seldomly concern the latency and model complexity \cite{nan2016delay,lau2016efficient}. Therefore, super-resolving online video sequences is still a challenging and necessary problem. In this paper, we will focus on the online VSR setting, where users need to receive super-resolved video sequences frame by frame in real time.

\begin{figure}[t!]
    \centering
    \includegraphics[width=1.0\linewidth]{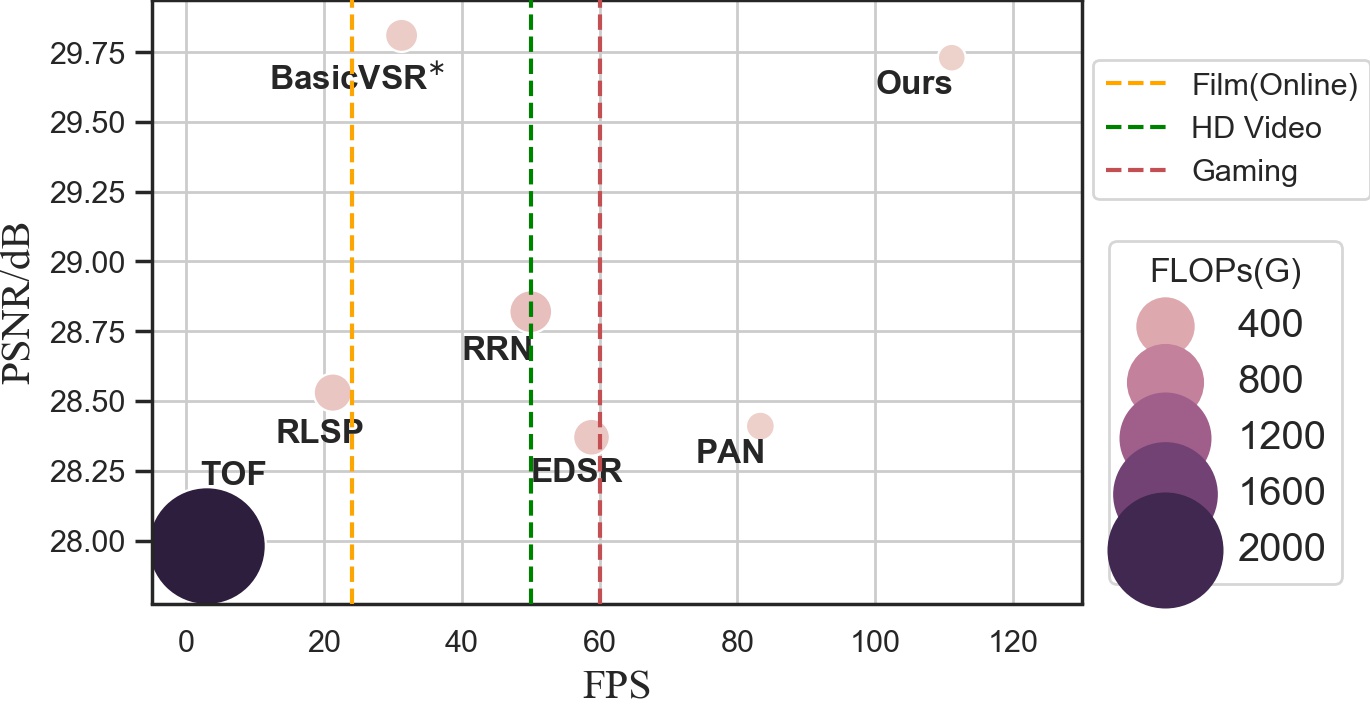}
    \caption{PSNR, FPS and FLOPs (G) of different methods deployed in Tesla V-100. The dot lines indicate the recommended FPS of different online applications, e.g., for ``\,Gaming\,'', the recommended FPS for cloud games is 60 FPS. ``\,BasicVSR$^{\ast}$\,'' is the modified version of BasicVSR to meet online VSR requirements. The input resolution is $180\times 320$.}
    \label{psnr_fps}
\end{figure}

In contrast to offline VSR, where there is no restriction on model complexity and latency, online VSR poses two key challenges to existing VSR methods. The first challenge is that online applications have extremely strict requirements on \emph{low latency and buffering lag}, because they involve real-time user interaction. In some online applications (e.g., online video conference), slight latency will significantly affect the user's experience and the conference quality, which is undesirable. However, most state-of-the-art VSR methods need to cache future frames for super-resolving the current frame, which will unavoidably introduce large latency. For example, TOF \cite{Xue2019} and RLSP \cite{fuoli2019efficient} use optical flows to align the future frames, and propagate the information contained in the future frames to the current frame for feature aggregation. Secondly, in order to reduce the transmission bandwidth, online VSR methods are usually deployed on client devices rather than cloud servers, which require the models to have low model complexity and real-time speed. However, the majority of deep VSR models adopt complicated network modules with high computational complexity in exchange for better performance, such as the progressive fusion blocks \cite{wang2019edvr,haris2019recurrent,caballero2017real}, the non-local attention blocks \cite{li2020mucan,yi2019progressive}, etc. The client devices (e.g., personal computers, mobile phones, etc.) are usually resources-constrained, which cannot support those models with such high computational complexity. It should be noted that the distortion quality of VSR methods is highly related to their model complexity. Reducing the model complexity will inevitably deteriorate the distortion quality of the generated images. Therefore, it is essential to achieve a good trade-off among the distortion quality of the generated videos, the processing latency, and model complexity for online-VSR algorithms.  

In this paper, we propose a low-latency online VSR solution based on a novel knowledge transfer-based method called convolutional kernel bypass graft (CKBG). To maximize the performance with no cached future frames and limited model capacity/latency, we propose to leverage the prior information from large pretrained image SR models. Borrowing the concept of heart bypass graft, CKBG enhances a VSR base model by bypassing the original network with ``kernel grafts'', which are extra convolutional kernels containing the prior knowledge of external pretrained image SR models. Specifically, given amounts of kernels extracted from large pretrained image SR models, CKBG learns a set of kernel bases by clustering in the Wasserstein space, and the ``kernel grafts'' are obtained by learning a linear representation under the space spanned by the learned bases. The grafted multi-branch structure of our network can be converted into a single-path structure with re-parameterization  \cite{ding2021diverse,ding2021repvgg}, which can effectively reduce the model complexity and latency.

The main contributions of this paper are as follows:
\begin{itemize}
\setlength{\itemsep}{0pt}
\setlength{\parskip}{0pt}
\setlength{\parsep}{0pt}
    \item In this paper, we mainly focus on online video super-resolution where no future frames are accessible and propose an extremely low-latency and effective online VSR method.  
    
    \item To further improve the performance, we propose the CKBG scheme, which incorporates the prior information learned from large pretrained image SR models into a VSR base model.
    
    \item  Experiment results show that our proposed method can process video sequences with up to 110 FPS and achieve promising performance, compared with other state-of-the-art deep VSR methods, as summarized in Figure \ref{psnr_fps}.
\end{itemize}

\section{Related Works}
\subsection{Deep Lightweight Image SR Methods}
In recent years, increasing efforts have been invested in exploring deep lightweight models for image SR because most promising deep SR methods \cite{lim2017enhanced,zhang2018image} require high computational complexity, which significantly limits their applications in resource-constrained devices. The common methods \cite{DRCN,DRRN,li2019feedback,li2019gated} for reducing model parameters adopt the recurrent structures, which share the weights and enhance the features with multiple cycles for image reconstruction. However, the recursive strategy increases the processing time and the performance gain is limited. Considering the effectiveness of the group convolution in \cite{howard2017mobilenets}, Ahn \etal \cite{ahn2018fast} proposed a cascaded convolutional network for image super-resolution. They combined the modified residual block with group convolution, significantly reducing the number of model parameters. The methods in \cite{hui2018fast,hui2019lightweight} extend the group convolution and propose an information-distillation block, which splits the input feature into several groups for further processing and then concatenates the output feature of each group for the feature fusion. These methods have shown remarkable trade-off performance in reducing model complexity and maintaining the distortion quality. With the split-and-concatenate strategy, Zhao \etal \cite{zhao2020efficient} proposed the pixel-attention mechanism to enhance useful information at each pixel location. Considering the characteristics of local regions, Xiao \etal \cite{xiao2021self} proposed an efficient method of generating dynamic convolution kernels, which adaptively extracts local features for image super-resolution. Unlike the above methods which design efficient modules by hands, the methods in \cite{chu2020multi,song2020efficient} employed the techniques of neural network architecture search to automatically find efficient model designs. Inspired by the structure re-parameterization techniques \cite{ding2021repvgg,ding2021diverse}, Zhang \etal \cite{zhang2021edge} proposed the edge convolutional block to accelerate the running speed of deep models. Even though these deep lightweight image SR models have shown their effectiveness, in terms of performance and real-time running speed, these methods do not consider the long-range temporal dependency when applied to video sequences. 

\subsection{Deep Video Super-resolution Methods}
With the rapid development of video applications, deep learning-based VSR methods have become increasingly popular. Unlike image SR, VSR needs to consider object motions and the temporal correspondence between successive frames. The method in \cite{caballero2017real} adopted the efficient spatial transformer for motion compensation, and then combine it with ESPCN \cite{shi2016real} to synthesize HR video sequences. As the object motions have the  property of being spatially variant, DUF \cite{jo2018deep} used a dual-path residual dense network to predict the residue between the ground-truth frames and the input LR frames in one path, and dynamically upsample the input LR frames in another path. In \cite{Xue2019}, it was found that incorporating an optical flow estimation network into a task-specific network for joint training, named TOF, is beneficial to the overall performance. However, its estimated motion field is different from the ground-truth optical flow, and the accuracy of the estimated optical flow is very sensitive to local illumination changes. Instead of using optical flow for the alignment of video frames, EDVR \cite{wang2019edvr} adopted deformable convolution \cite{dai2017deformable} to align the features from multi-scale levels. According to \cite{lin2021fdan}, deformable convolution cannot effectively capture the long-range dependency and suffer from unstable training. In practice, the running speed of deformable convolution is very slow, which cannot satisfy the real-time requirements. To achieve better efficiency, Fuoli D \etal \cite{fuoli2019efficient} proposed an efficient recurrent network to investigate the information from adjacent frames only for practical VSR tasks. This method directly feeds the extracted features of the hidden state from the previous step into the current step for feature fusion so that information can temporally propagate along the video sequences. Since VSR is an ill-posed problem, RSDN \cite{isobe2020video} introduced structural and detailed information to regularize the process for generating HR frames. Recently, bi-directional recurrent methods \cite{chan2021basicvsr,huang2017video,chan2021basicvsr,chan2021basicvsr++}, such as BasicVSR and its variant \cite{chan2021basicvsr,chan2021basicvsr++}, have shown their effectiveness in VSR, which can fully exploit the information from the forward and the backward directions of input video sequences. However, these methods need to acquire the whole video sequences beforehand and hence, they are impractical in online-VSR scenarios. It should be noted that video sequences are transmitted in the streaming format in online applications. Therefore, it is impractical to capture the information contained in the future frames because undesirable latency is introduced. In this paper, we will focus on the online VSR scenarios in which future frames are inaccessible and the deployed devices are resource-constrained, such as low-configuration devices.

\subsection{Knowledge Transfer for Image and Video Super-resolution}
Knowledge distillation is a well-known technique to transfer knowledge from a large deep model to a smaller one. In knowledge distillation, the large model is called the teacher network, while the smaller model is called the student network. Recently, Gao \etal \cite{gao2018image} proposed to distill knowledge from a teacher image super-resolution model by minimizing the distance of statistical properties (i.e., maximum values, mean values, etc.) of feature maps between the teacher network and the student network. He \etal \cite{he2020fakd} proposed a feature-affinity distillation (DAKD) method for image super-resolution, which transfers knowledge by using the correlation matrices of feature maps. Lee \etal \cite{lee2020learning} proposed to leverage the privileged information from ground-truth images and distill the knowledge by minimizing the distance between the features of the teacher network and the student network. Xiao \etal \cite{xiao2021space} proposed an effective knowledge-distillation method for video super-resolution, which enforces the spatial and temporal characteristics of the teacher network and student network to be consistent. All these distillation-based super-resolution models require the network topology of the teacher network and the student network to be consistent. In contrast, our proposed CKBG does not restrict the network structures, which provides more flexibility than distillation-based methods. In addition, the large teacher models used in our CKBG-based model are not involved in the training process, which is significantly different from the distillation-based methods.

\section{The Proposed Method}
\begin{figure}[t!]
    \centering
    \includegraphics[width=1.0\linewidth]{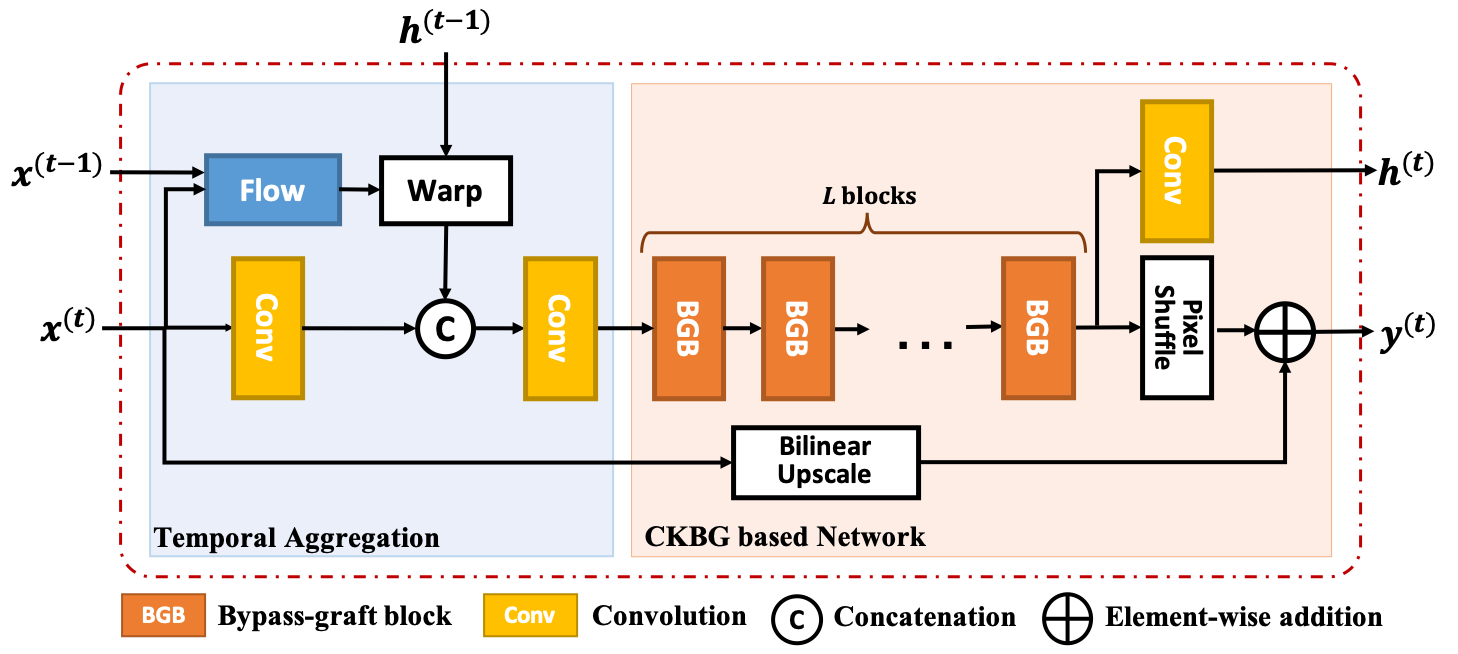}
    \caption{The overall pipeline of our proposed method.}
    \label{network_structure}
\end{figure}

The overall structure of our proposed method is illustrated in Figure \ref{network_structure}. Our method adopts the recurrent structure, which only utilizes the information in the current and previous frame. Therefore, it does not need the extra cost of caching frames in the super-resolution process. In particular, the proposed network first uses optical flow to align the features extracted from the previous frame with the current frame and then, temporally fuse the aligned features with the features extracted from the current frame. After that, the temporally aggregated feature is forwarded to the cascaded bypass-graft blocks (BGBs) for feature extraction, constructed based on our proposed CKBG method. At the output of our model, we adopt the PixelShuffle operator to upsample the extracted features and combine them with the bilinearly upscaled LR inputs for generating the final HR frames.

\subsection{Temporal Aggregation}
For temporal aggregation, we first utilize SpyNet \cite{ranjan2017optical}, denoted as $S(\cdot)$, to estimate the optical flow $f^{(t-1)}$  between the current LR frame $x^{(t)}$ and the previous frame $x^{(t-1)}$, which is computed as follows:
\begin{equation}
    f^{(t-1)} = S(x^{(t)}, x^{(t-1)}).
\end{equation}
Then, we use the estimated optical flow to perform alignment in the feature space. The warped feature of the previous frame is computed as follows:
\begin{equation}
        \hat{h}^{(t-1)} = \text{Warp}(h^{(t-1)}, f^{(t-1)}),
\end{equation}
where $h^{(t-1)}$ represents the feature extracted from the previous frame, $\hat{h}^{(t-1)}$ is the corresponding warped feature, and $\text{Warp}(\cdot)$ represents the warp operator. To avoid significantly increasing the model complexity, we simply concatenate the feature extracted from the current frame with the aligned feature along the channel dimension. Then, a convolutional layer is used to aggregate the features temporally. The aggregated feature $F^{(t)}$ is calculated as follows:
\begin{equation}
        F^{(t)} =\text{conv}(\text{Cat}(\text{conv}(x^{(t)}), \hat{h}^{(t-1)})),
\end{equation}
where $\text{Cat}(\cdot)$ is the concatenation operator and $\text{conv}(\cdot)$ represents a $3\times 3$ convolutional operator.

\begin{figure*}[t]
    \centering
    \includegraphics[width=\linewidth]{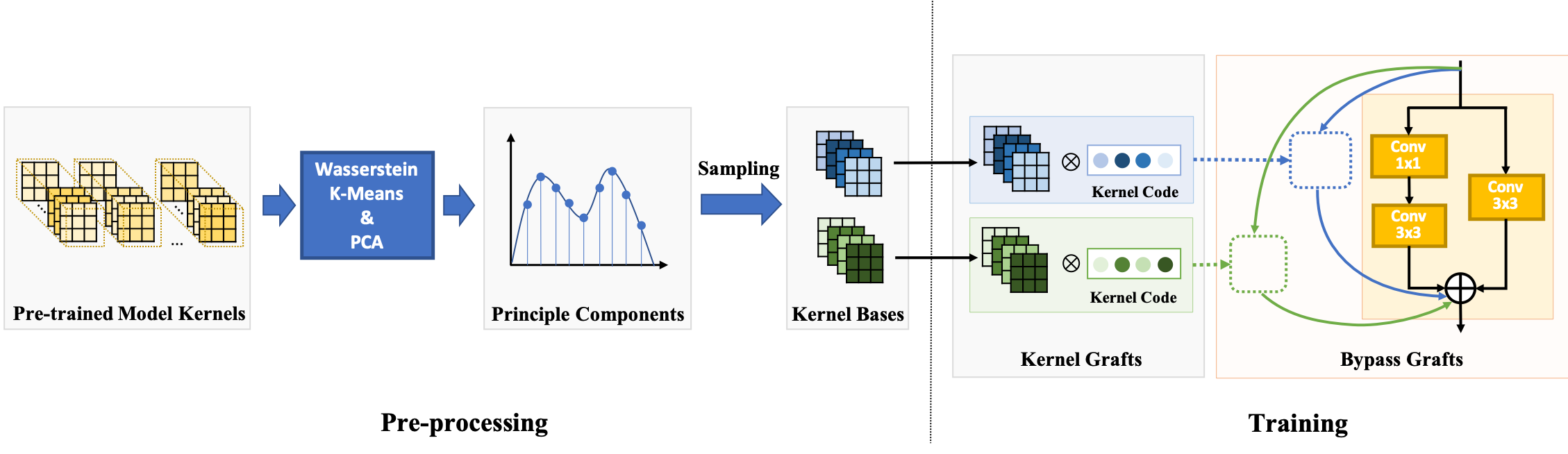}
    \caption{The overall pipeline of the kernel bypass graft in the preprocessing and training stages. In the testing phase, all the parallel convolutional kernels are merged into a single convolutional kernel. Note: the kernel bases are fixed in the training phase.}
    \label{kernel_graft}
\end{figure*}

\subsection{Convolutional Kernel Bypass Graft}
\subsubsection{Wasserstein Distance}
Suppose $\mu$ and $\nu$ are probability measure defined on the Polish spaces $\mathcal{X}$ and $\mathcal{Y}$, respectively. Let $\mathcal{P}(\mathcal{X}\times \mathcal{Y})$ be a set of probability distributions defined over the space $\mathcal{X}\times \mathcal{Y}$. The joint distribution of $\mu$ and $\nu$ is denoted as $\prod(\mu, \nu)$, i.e., $\prod(\mu, \nu)=\{\pi\in \mathcal{P}(\mathcal{X}\times \mathcal{Y})\vert \int_{\mathcal{X}}\mathrm{d}\pi(x,y)=\mathrm{d}\nu(y), \int_{\mathcal{Y}}\mathrm{d}\pi(x,y)=\mathrm{d}\mu(x)\}$, where $x\in\mathcal{X}, y\in\mathcal{Y}$, and $\pi$ is the joint probability distribution. The optimal transportation problem aims to find an optimal transport mapping with the minimum transportation cost between two locations $x$ and $y$, which is defined as follows:
\begin{equation}
    \min_{\pi\in \prod(\mu,\nu)}\int_{\mathcal{X}\times \mathcal{Y}}c(x,y)\,\mathrm{d}\pi(x,y),
\end{equation}
where $c(x,y):\mathcal{X}\times\mathcal{Y}\rightarrow \mathbb{R}^{+}\cup \{0\}$ denotes the cost function from $x$ to $y$. This is the well-known Kantorovich's formulation. In this case, $\pi$ is also called a transport mapping, and $\mathrm{d}\pi(x,y)$ specifies the transported mass between $x$ and $y$. Based on this, the Wasserstein distance \cite{villani2009optimal} is defined as follows:
\begin{equation}
    \mathcal{W}_{p}^{p}(x,y) = \left(\min_{\pi\in \prod(\mu,\nu)} \int_{\mathcal{X}\times \mathcal{Y}}\vert x - y\vert^{p}\,\mathrm{d}\pi(x,y)\right)^{1/p}, \label{op}
\end{equation}
where $c(x, y)=\vert x-y\vert^{p}$. If $p=2$, it is called the $2$-Wasserstein distance, denoted as $\mathcal{W}_{2}^{2}$. Assuming that there are $N_{1}$ probability distributions $\{\mu_{i}\}_{i=1}^{N_{1}}$ defined in the $2$-Wasserstein metric space, the Fr\'echet mean of these distributions is defined as follows:
\begin{equation}
        \nu  = \mathop{\arg\min}_{\nu \in\mathcal{P}(\mathcal{Y})}\sum_{i=1}^{N_{1}}\lambda_{i}\mathcal{W}_{2}^{2}(\mu_{i}, \nu), \label{WB}
\end{equation}
where $\lambda_{i}$ is the weight associated with the $i$-th probability distribution, and $\sum_{i=1}^{N_{1}}\lambda_{i}=1$. The solution $\nu$ is also called the Wasserstein barycenter, which achieves the minimum weighted $2$-Wasserstein distance for every $\mu_{i}$.

\subsubsection{Kernel Prior Learning and Grafting}
As shown in Figure 3, the proposed CKBG is based on a two-stage learning framework. At the first stage, CKBG extracts convolutional kernels from a pretrained image SR model, and then learns a set of kernel bases by clustering the extracted kernels in the Wasserstein metric space. After that, the ``kernel grafts'' are obtained by learning linear representations under the space spanned by the kernel bases. A ``kernel graft'' is grafted alongside each convolutional layer in the original network, forming a bypass-graft block (BGB). It should be noted that the obtained BGB has a multi-branch structure in the training stage, but the multi-branch structure can be further converted into a single-path structure for acceleration in the testing phase. In this section, we will elaborate on the learning, grafting, and testing of the proposed CKBG.

\begin{figure*}[bt!]
\centering
\subfloat[]{
        \includegraphics[width=0.67\linewidth]{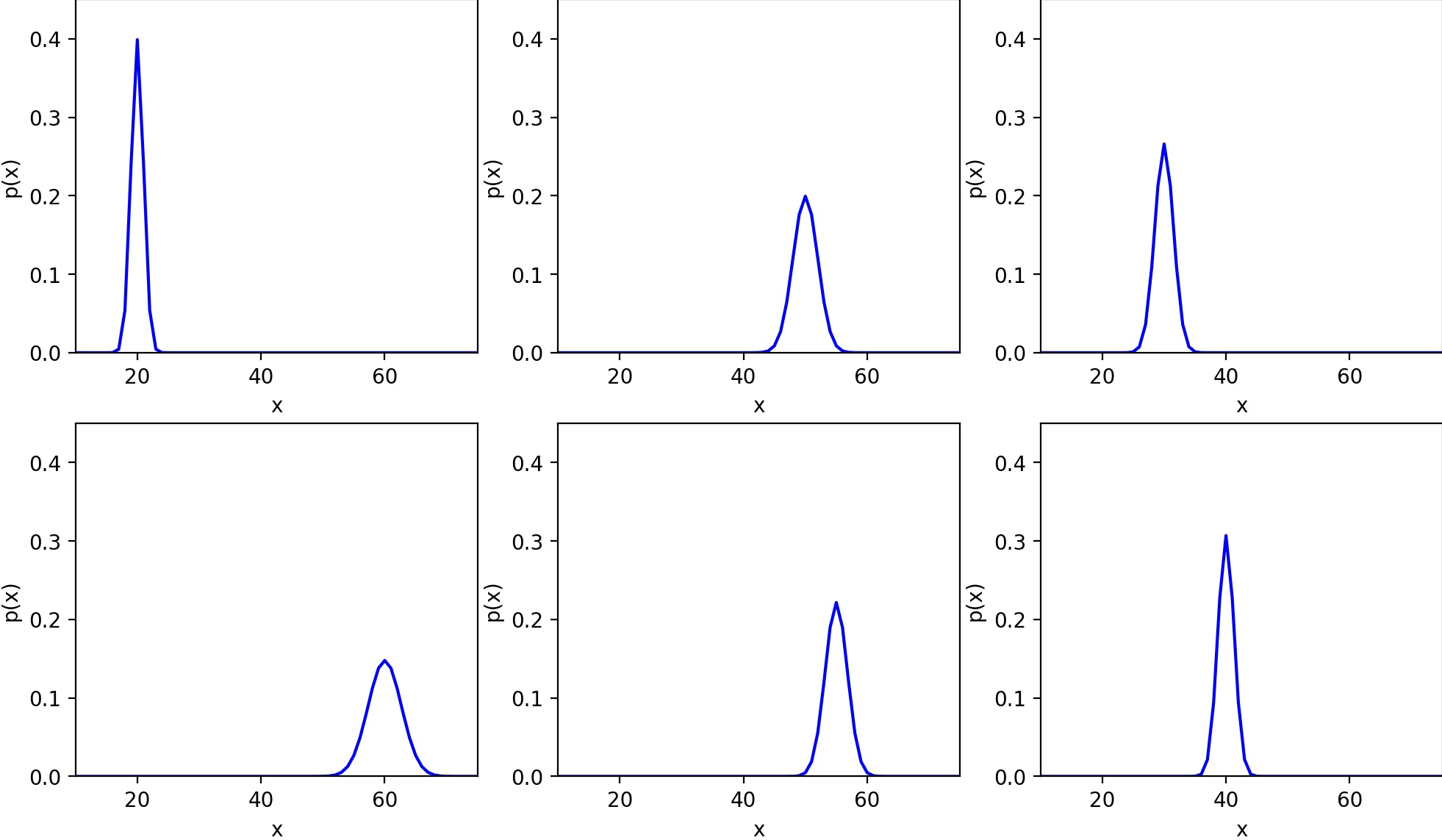}
    } 
\subfloat[]{
        \includegraphics[width=0.26\linewidth]{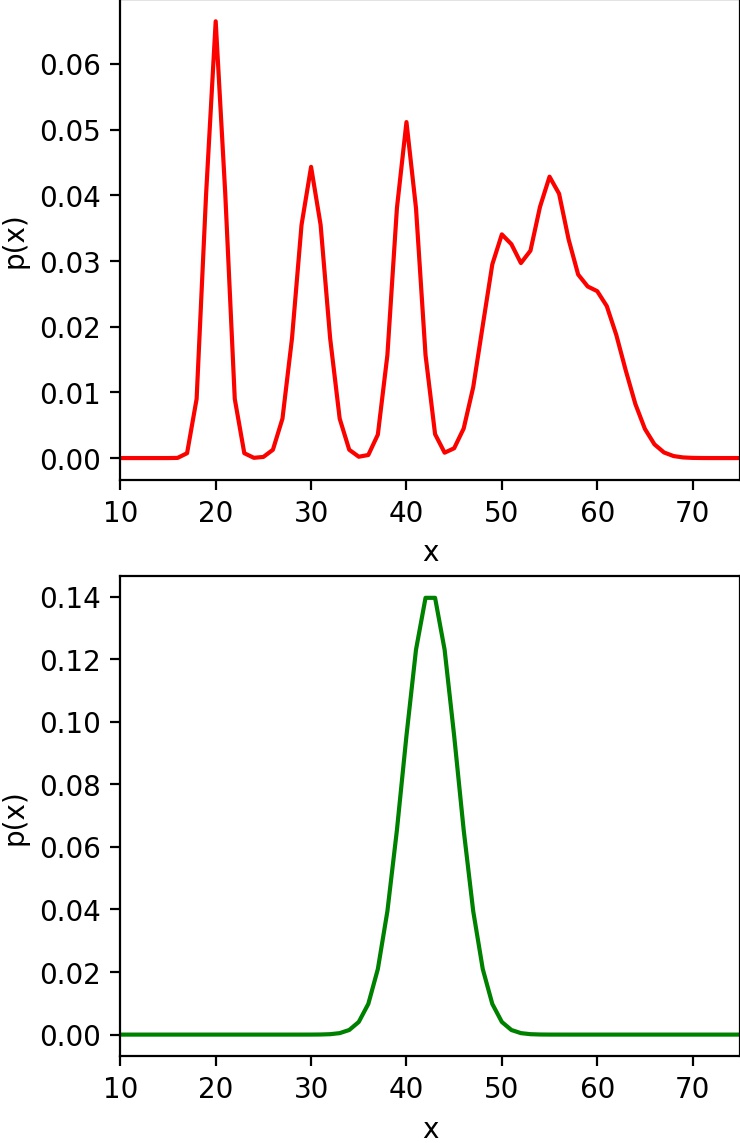}
    }
    \caption{(a). Samples of different unimodal distributions. (b). The centroids are computed in the Euclidean space (\textcolor{red}{red}) and the Wasserstein space (\textcolor{green}{green}). The horizontal axis denotes the support of the distributions, and the vertical axis is the value of the distributions. }
    \label{1-d}
\end{figure*}

\textbf{Kernel Prior Learning}. Given a pretrained image SR model, we first extract convolutional kernels from it. In this paper, the adopted image SR model is EDSR \cite{lim2017enhanced}, which is a classic image SR model and has shown promising results. The extracted kernels contain a lot of prior information for generating HR images. 
To remove redundancy and obtain more representative kernels, we cluster similar kernels by performing the K-Means algorithm in the Wasserstein space, resulting in a set of cluster centroids. Specifically, assuming that $K$ convolutional kernels are obtained from a pretrained image SR model, denoted as $\{\mathbf{k}_{i}\}_{i=1}^{K}$, the cluster centroids are obtained by solving the following optimization problem:

\begin{equation}
\mathop{\arg\min}_{\mathbf{S}}\sum_{i=1}^{M}\sum_{\mathbf{k}\in S_{i}}\mathcal{W}_{2}^{2}(\mathbf{k}, \mathbf{c}_{i}), \label{kwb}
\end{equation}
where $\mathbf{S}=\{S_{1},\cdots, S_{M}\}$ denotes the partition of the extracted kernels and $M$ is the number of clusters. $\mathbf{c}_{i}$ represents the $i$-th cluster centroid in the Wasserstein space. From Eqn.\,(\ref{WB}), we can find that obtaining the kernel centroids is equivalent to computing the Wasserstein barycenters for each kernel cluster. In this case, the weights associated with the Wasserstein distance between the kernels and the centroids are all the same. We use the variational method provided in \cite{mi2018variational} to solve the above optimization problem. To better inherit the prior information from the extracted kernels, we perform K-Means clustering in the Wasserstein space rather than in the Euclidean space, because the optimal transport mapping can better preserve the geometric properties (e.g., shape) of the extracted kernels. As a result, the learned cluster centroids learned in the Wasserstein space have a similar geometric structure to the extracted kernels. Figure \ref{1-d}(a) illustrates a $1$-D example of six unimodal distributions. As shown in Figure \ref{1-d}(b), we can find that the centroid distribution computed in the Euclidean space is severely distorted. In contrast, the geometric properties of the cluster centroids obtained in the Wasserstein space are more similar to the original distributions. It is worth noting that the output responses are highly related to the geometric properties of the kernels, so the kernel centroids obtained in the Wasserstein space can avoid generating distortion, leading to similar output responses.


The proposed CKBG enhances the base VSR model by learning a set of new kernels and grafting the kernels in the original network. We called the learned kernels ``kernel grafts''. 
In this paper, the ``kernel grafts'' are learned based on a kernel space constructed from the cluster centroids, which contain prior knowledge extracted from the pretrained image SR model. Specifically, this kernel space is represented by the kernel bases obtained by performing principal component analysis on the cluster centroids, as follows:
\begin{equation}
    CC^{T}=U\Sigma U^{T},
\end{equation}
where $C=[c_{1}, \cdots, c_{M}]$ is the cluster centroid matrix, $C^{T}$ is its transpose matrix, $U=[\mathbf{u}_{1}, \cdots, \mathbf{u}_{M}]$ is the eigenvector matrix of $C$, and $\Sigma=\text{Diag}(\sigma_{1},\cdots, \sigma_{M})$ contains the corresponding eigenvalues sorted in the descent order. The eigenvectors in $U$ are the principal components of the cluster centroids, which can be used as the potential bases of the kernel prior sub-space. Larger singular values indicate that the corresponding eigenvectors are more significant.

\textbf{Kernel Graft}. As shown in Figure \ref{kernel_graft}, for each convolutional layer in the base network, several kernel grafts are obtained, which bypass the original convolutional kernels. 
For each kernel graft, we first select a set of bases to form the kernel space, and the selected bases are the principal components of the cluster centroids sampled from the following  categorical distribution:
\begin{equation}
    \mu_{i} \sim p(\mathbf{\Theta}),
\end{equation}
where $\mathbf{\Theta}=[\theta_{1}, \cdots, \theta_{M}]$, $\theta_{i} = \sigma_{i}/\sum_{j=1}^{M}\sigma_{j}, \forall i=1,\cdots, M$. 
This implies that the bases are sampled according to their significance, which corresponds to their singular values. The more significant principal components are more likely to be selected as the bases for the kernel space.
Meanwhile, the randomness introduced by the sampling process brings diversity of kernel grafts. 

\begin{figure}[tb!]
    \centering
    \includegraphics[width=0.95\linewidth]{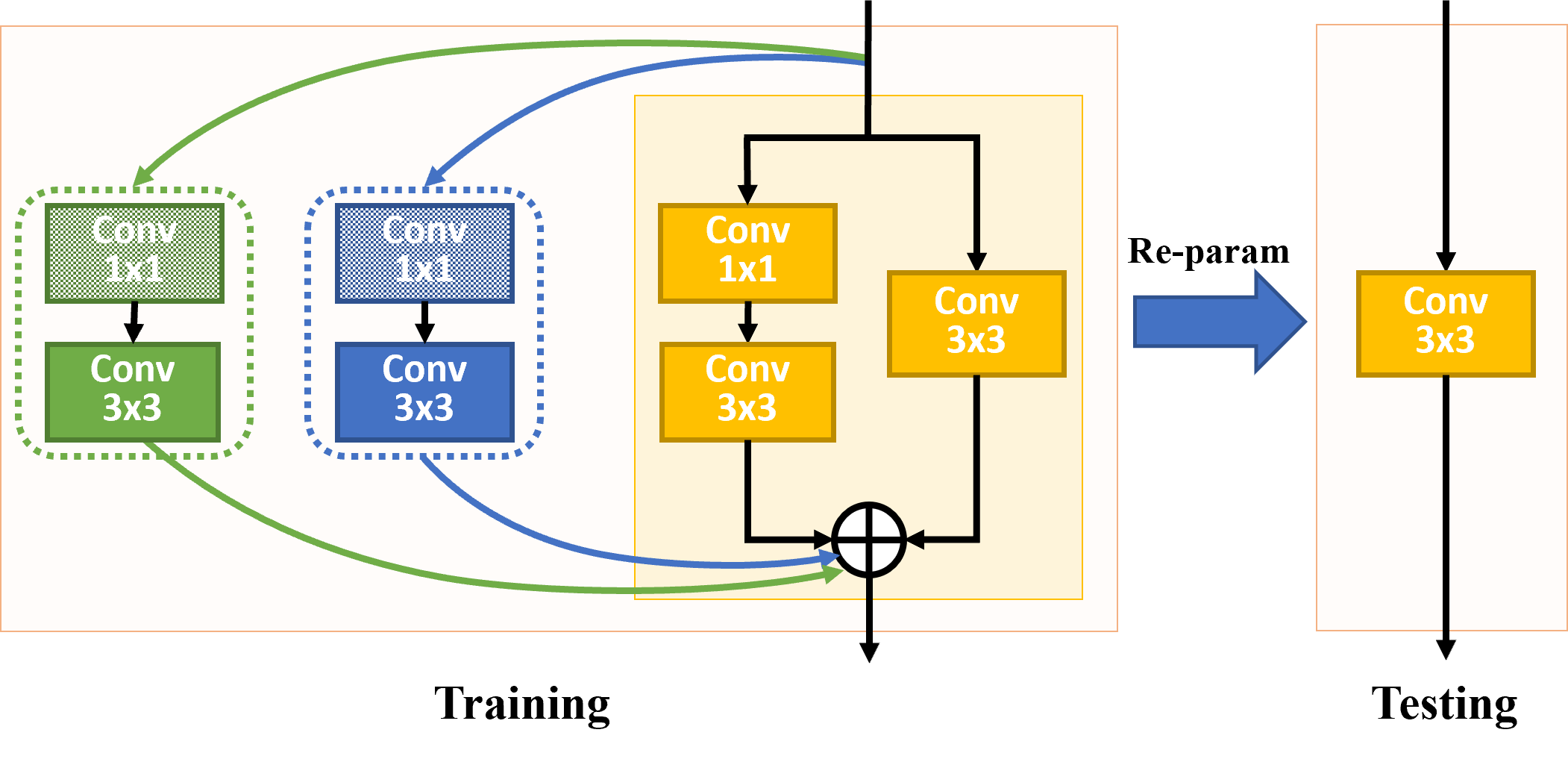}
    \caption{The overall structure of the proposed bypass-graft block (BGB) in the training stage and the testing stage. The \textcolor{green}{green} and \textcolor{blue}{blue} arrows denote the grafted branches. Note: the grafted kernels are fixed during training. }
    \label{wp-conv}
\end{figure}

Finally, the convolutional kernel graft is obtained by learning a linear combination of the sampled kernel bases. As shown in Figure \ref{wp-conv}, in our implementation, learning a linear combination of the kernel bases is equivalent to adding a $1\times 1$ convolution (i.e. the patterned boxes) before the ``kernel graft'' , which is updated simultaneously with the parameters of the base network, while the parameters of ``kernel grafts'' are kept fixed.

\textbf{BGB in testing}. In the testing phase, the multi-branch structure of BGB can be converted into a single convolution according to the linear property of the convolutional operation \cite{ding2021diverse}. The network structure of the proposed BGB in the training and testing stages is illustrated in Figure \ref{wp-conv}. Specifically, the consecutive $1\times 1$ and $3\times 3$ convolutional operators are first merged to form a single convolution. Then, all the parallel $3\times 3$ convolutional operators are merged into a single convolutional operator, resulting in a highly efficient  single-path structure. It is worth noting that the re-parameterized kernel at the testing stage is equivalent to the original multi-branch structure in the training stage, without sacrificing any performance. More details about kernel re-parameterization can be found in Appendix A. 

\subsection{Loss function}
We adopt the Charbonnier loss to measure the distance between the generated frames and the ground-truth (GT) frames, which is defined as follows:
\begin{equation}
    L = \frac{1}{T}\sum_{t=1}^{T}\sqrt{\left\Arrowvert I_{SR}^{(t)}-I_{GT}^{(t)}\right\Arrowvert_{2}^{2}+\epsilon},
\end{equation}
where $I_{SR}^{(t)}$ and $I_{GT}^{(t)}$ denote the generated SR frame and the ground-truth frame at the $t$-th time step, respectively, $T$ is the number of input LR frames, and $\epsilon$ is a hyper-parameter.

\section{Experiments and Analysis}
\textbf{Datasets}. The REDS dataset \cite{Son_2021_CVPR} and Vimeo-90K dataset \cite{Xue2019}, which are two widely used datasets in VSR, are adopted for training. In the testing stage, the REDS4 dataset \cite{chan2021basicvsr} and Vid4 dataset \cite{liu2013bayesian} are used for evaluation. In these two datasets, the upscaling factor is 4, and the bicubic downsampling kernel is used to generate LR video sequences.

\textbf{Evaluation Metrics}. The performance of online VSR algorithms should include three perspectives: model complexity, latency, and the distortion quality of the generated videos. We measure model complexity by considering the number of model parameters, floating point operations per second (FLOPs), and the number of activations. For distortion quality assessment, peak signal-to-noise ratio (PSNR) and structural similarity index measure (SSIM) are adopted for evaluation. Since our proposed method focuses on online scenarios, the latency $t$ is very important and is defined as follows:
\begin{equation}
    t = t_{cache} + t_{run},
\end{equation}
where $t_{cache}$ denotes the time to cache the frames required by a VSR method and $t_{run}$ denotes the runtime required to super-resolve an input frame. If the frame rate is 24 FPS, then the cache time for one frame is 40\,ms. It is worth noting that distortion and latency are well-known trade-offs. The trade-off score function \cite{ignatov2021real} is used to objectively measure the performance of the algorithms in online situations and is defined as follows:
\begin{equation}
    \text{score} = \frac{2^{\text{PSNR}}}{C \times t},
\end{equation}
where $C$ is a constant and set to $2^{20.0}$ in the experiment. A model with a higher score can achieve a better trade-off between distortion quality and latency, so the model is more effective and applicable in online scenarios.

\textbf{Implement Details}. In the experiments, we train our proposed model using patches of size $80\times 80$ randomly cropped from the input video sequences. For data augmentation, we randomly flipped and rotated the input video sequences. The number of channels in our model is set to $64$, and the batch size is $8$ in the training process. We use Adam \cite{kingma2014adam} with $\beta_{1}=0.9$ and $\beta_{2}=0.999$ to update the weights of the model. The initial learning rate is $2\times 10^{-4}$ and the Cosine annealing strategy is utilized to adaptively adjust the learning rate in the training process. The total number of iterations used in training is $6\times 10^{5}$. We implemented the proposed method with PyTorch in Tesla V-100 GPUs, and it took approximately eight days to complete the training. 
\begin{table*}[!t]
\centering
\caption{Comparison of the average PSNR, SSIM, model complexity, latency, and trade-off score of different VSR methods. Note that  ``\,\# Param.\,'' and ``\,\# Act.\,'' represent the number of model parameters and the number of activations, respectively.   ``\,Online\,'' refers to whether a method can be applied online. ``\, Time\,'' represents latency measured on the REDS4 dataset in Tesla-V100. The average size of the input frame is $180\times 320$.  `` RGB '' and `` Y '' mean that the evaluation metrics are measured in the RGB space and Y channel, respectively. The best results are highlighted in bold. The second-best results are underlined.}
\begin{tabular}{|cc|c|c|c|c|c|c|c|c|}
\hline
\multicolumn{2}{|c|}{\multirow{2}{*}{Methods}} & \multirow{2}{*}{\# Param.} & \multirow{2}{*}{FLOPs} & \multirow{2}{*}{\# Act.} &\multirow{2}{*}{Online} & \multirow{2}{*}{Time} & \multirow{2}{*}{Score} & REDS4 (RGB) & Vid4 (Y) \\ \cline{9-10} 
\multicolumn{2}{|c|}{}                  &                   &                   &                   &   &                &                   & PSNR/SSIM    & PSNR/SSIM \\ \hline
\multicolumn{1}{|c|}{} & Bicubic           &        -          &        -          &        -          &    &   -           &       -           & 26.13/0.7388 & 23.78/0.6347 \\ \hline 
\multicolumn{1}{|c|}{\multirow{2}{*}{SISR}} &EDSR-M \cite{lim2017enhanced}  &     1,571\,K             &   114.28\,G       &  201.83\,M &\cmark        &  17\,ms           & 19.46  & 28.37/0.8078 & 25.31/0.6608 \\  
\multicolumn{1}{|c|}{} & PAN \cite{zhao2020efficient}   &     \textbf{272\,K}      &  \underline{28.29\,G}       &  237.88\,M  &\cmark        &  \underline{12\,ms}           &  \underline{28.34} &  28.41/0.8089 & 25.35/0.7306 \\ \hline
\multicolumn{1}{|c|}{\multirow{5}{*}{VSR}} & TOF\cite{Xue2019} &    1,405\,K       &   2,175.25\,G     &   1,251.93\,M    &\xmark     &      334\,ms      &     0.75           & 27.98/0.7990 & 25.89/0.7651 \\ 
\multicolumn{1}{|c|}{} & RLSP \cite{fuoli2019efficient}  &  \underline{1,154\,K}   &  132.94\,G        &  \underline{108.74\,M}  &\xmark       &    47\,ms       &   7.86      & 28.53/0.8136  & 25.69/0.7530 \\ 
\multicolumn{1}{|c|}{} &RRN \cite{isobe2020revisiting}  &  3,364\,K             &  193.62\,G          &  164.96\,M   &\cmark     &   20\,ms        &      22.59     & 28.82/0.8234 & 25.85/0.7660 \\ 
\multicolumn{1}{|c|}{} & BasicVSR$^{\ast}$ \cite{chan2021basicvsr}    &  1,887\,K        &    71.33\,G  & 185.24\,M  &\cmark     &   32\,ms        &    27.38      & \textbf{29.81/0.8535}  & \textbf{26.47}/0.7986 \\ 
\multicolumn{1}{|c|}{} & CKBG(ours)  &    1,750\,K       &    \textbf{17.85\,G}               &  \textbf{34.09\,M}   &\cmark              &      \textbf{9\,ms}          &      \textbf{91.14}             & \underline{29.73/0.8514} & \underline{26.34/0.7857} \\ \hline
\end{tabular}
\label{exp1}
\end{table*}
\begin{figure*}[!t]
    \centering
    \includegraphics[width=0.95\linewidth]{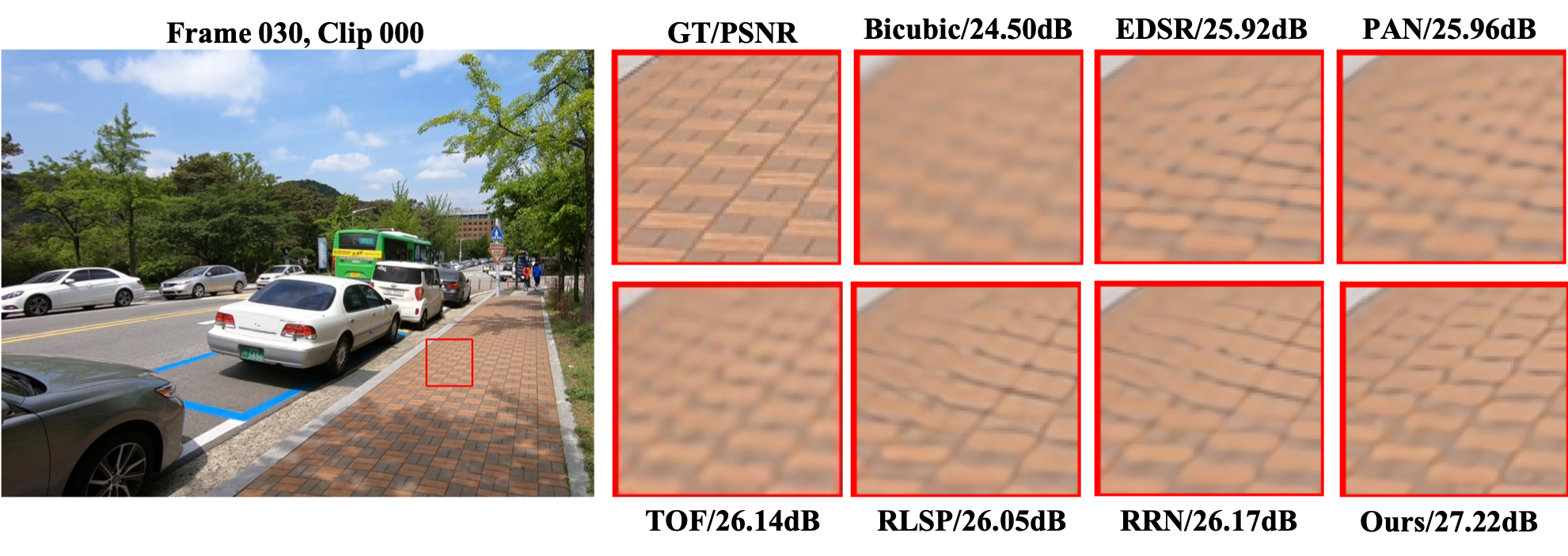}
    \caption{The illustrated image is selected from the REDS4 dataset. The region marked by the red box is generated by different VSR methods for visual comparison. }
\label{visual_results1}
\end{figure*}

\begin{figure*}[!t]
    \centering
    \includegraphics[width=0.95\linewidth]{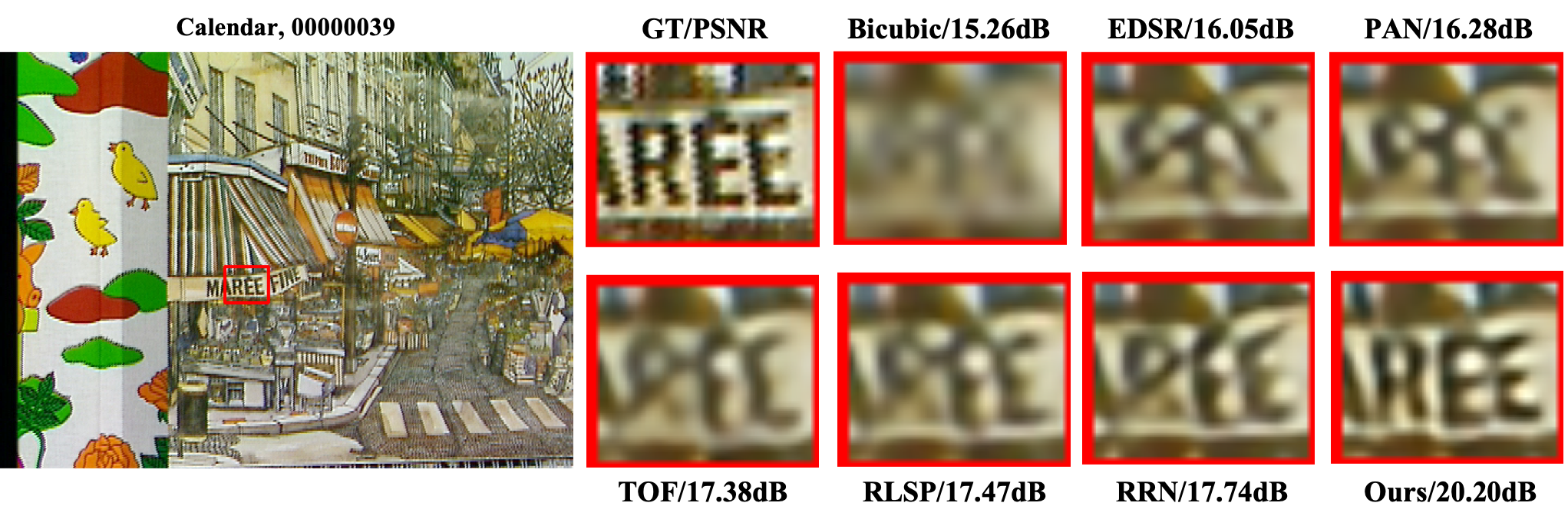}
    \caption{The illustrated image is selected from the Vid4 dataset. The region marked by the red box are generated by different VSR methods for visual comparison. }
\label{visual_results2}
\end{figure*}
\subsection{Experiment on REDS4 and Vid4 Datasets}
In this experiment, we compared our proposed method with EDSR-M \cite{lim2017enhanced}, PAN \cite{zhao2020efficient}, RLSP \cite{fuoli2019efficient}, RRN \cite{isobe2020revisiting}, TOF \cite{Xue2019}, and BasicVSR \cite{chan2021basicvsr}. Among these methods, EDSR-M and PAN are single-image SR models with lower model complexity. Since online video SR algorithms have a high requirement in latency, these two methods do not need to cache frames and have the ability to process video in real time. RLSP and RRN are recurrent-based video SR methods and have achieved promising results. Specifically, RRN only incorporates the information of the previous frame into the current state for generating HR frames, so this method is also suitable for online scenarios. Both methods were originally trained on videos blurred by Gaussian kernels, so we retrained these models with the same training datasets for a fair comparison. BasicVSR is a bi-directional VSR model, so it should not be applicable for online scenarios. We modified BasicVSR by removing the backward part and reducing its model size for online application, which is denoted as BasicVSR$^{\ast}$. The implementation details of BasicVSR$^{\ast}$ can be found in Appendix B. We used the open-source codes provided by the authors to implement the compared methods. All the evaluation results, including the number of model parameters, the number of activations,  FLOPs, the average PSNR and SSIM, the required runtime, and the trade-off score, are reported in Table \ref{exp1}. 

As observed, our proposed model has the lowest model computational complexity because our proposed method requires less FLOPs than other compared methods, including two image SR methods, EDSR-M and PAN. Although our proposed method uses an external network to estimate optical flow, the number of model parameters required by our method is slightly higher than that of EDSR-M, which is acceptable. For distortion quality, our proposed method significantly outperforms EDSR-M, PAN, TOF, RLSP, and RRN in terms of PSNR and SSIM. This shows that our proposed method is more efficient and effective for VSR. Surprisingly, we found that the latency of our proposed method measured in Tesla V-100 is less than 10ms on the REDS4 dataset. The average size of video sequences in the RED4 dataset is $180\times 320$, and the upscaling factor used in our experiments is 4. This means that our proposed method takes less than 10ms to produce a video frame of the size $720\times 1280$ in online scenarios, which is significantly efficient. In addition, our method has much lower latency than the most efficient image SR model (i.e., PAN) and runs much faster than other compared VSR methods. Overall, compared to other efficient image SR and VSR methods, our proposed method requires extremely low latency and achieves competitive performance in terms of distortion quality. Therefore, our proposed method has a higher trade-off score, which shows that our proposed method outperforms other compared methods in online situations.

Figures \ref{visual_results1} and \ref{visual_results2} illustrate the visual results of different VSR methods. For better visualization, a region of each image marked by a red rectangular box is cropped and enlarged. In Figure \ref{visual_results1}, we can find that the video frames generated by EDSR, PAN, RLSP, TOF, and RRN suffer from severe distortion and lack of sharp texture information in the marked dense grids, while our method can effectively preserve object details (e.g., edges and textures). Similarly, in Figure \ref{visual_results2}, we can easily observe that our proposed method has a better ability to generate clear image content (e.g., the word `` REE'' in the marked region) than other compared methods. Overall, the illustrated results show that our proposed method has a better ability in producing images with low distortion and high visual quality. Please find more visual results in Appendix C.

\subsection{Experiment on Different Hardware Devices}
Online VSR methods are usually deployed in clients' devices, and the configurations of personal devices are various. In this experiment, we evaluated the latency of different VSR methods in two devices, NAVIDIA 2080 Ti and 1080 Ti. These two devices have been widely used in gaming applications, which require video algorithms to process up to 60 FPS\footnote{https://www.gamingscan.com/best-fps-gaming}. The REDS4 dataset is used in this experiment, where the size of the generated video frame is $720\times 1280$. The evaluation results on 2080 Ti and 1080 Ti are shown in Table \ref{table_time}. In addition, the relationship between the PSNR and the FPS achieved is illustrated in Figure \ref{psnr_fps}. We can find that the latency of our method is significantly lower than other methods. In particular, for 2080 Ti, the processing speed of our proposed method is up to 110 FPS. It is worth noting that none of three compared methods can meet the latency requirements of online gaming on the devices equipped with 1080 Ti, but our method can amazingly process the video sequences at a rate of up to 80 FPS. It is worthy noticing that personal laptops equipped with 1080 Ti are not high-configuration for online games. This shows that our method is less affected by hardware devices and more applicable to different industrial products.

\begin{table}[htbp]
\centering
\caption{The latency required by different methods running on various devices. All models are evaluated on the REDS4 dataset, and the average size of the input frame is $180\times 320$. The best results are highlighted in bold.}
\begin{tabular}{|cc|cccc|}
\hline
\multicolumn{2}{|c|}{\multirow{2}{*}{}}    & \multicolumn{4}{c|}{Methods}                                                    \\ \cline{3-6} 
\multicolumn{2}{|c|}{}                     & \multicolumn{1}{c|}{EDSR-M} & \multicolumn{1}{c|}{PAN} & \multicolumn{1}{c|}{RRN} & CKBG(ours) \\ \hline
\multicolumn{2}{|c|}{\#Param}                     & \multicolumn{1}{c|}{1,571\,K} & \multicolumn{1}{c|}{\textbf{272\,K}} & \multicolumn{1}{c|}{3,364\,K} & 1,750\,K  \\ \hline
\multicolumn{2}{|c|}{PSNR}                     & \multicolumn{1}{c|}{28.37} & \multicolumn{1}{c|}{28.41} & \multicolumn{1}{c|}{28.82} & \textbf{29.73}  \\ \hline
\multicolumn{1}{|c|}{\multirow{3}{*}{2080Ti}} & Time & \multicolumn{1}{c|}{22\,ms} & \multicolumn{1}{c|}{16\,ms} & \multicolumn{1}{c|}{27\,ms} & \textbf{9\,ms} \\ \cline{2-6} 
\multicolumn{1}{|c|}{}                  & FPS & \multicolumn{1}{c|}{45.45} & \multicolumn{1}{c|}{62.5} & \multicolumn{1}{c|}{37.03} & \textbf{111.11} \\ \cline{2-6} 
\multicolumn{1}{|c|}{}                  & Gaming & \multicolumn{1}{c|}{\xmark} & \multicolumn{1}{c|}{\cmark} & \multicolumn{1}{c|}{\xmark} & \cmark  \\ \hline
\multicolumn{1}{|c|}{\multirow{3}{*}{1080Ti}} & Time & \multicolumn{1}{c|}{40\,ms} & \multicolumn{1}{c|}{36\,ms} & \multicolumn{1}{c|}{44\,ms} & \textbf{12\,ms} \\ \cline{2-6} 
\multicolumn{1}{|c|}{}                  & FPS & \multicolumn{1}{c|}{25.00} & \multicolumn{1}{c|}{27.77} & \multicolumn{1}{c|}{22.72} & \textbf{83.33}  \\ \cline{2-6} 
\multicolumn{1}{|c|}{}                  & Gaming & \multicolumn{1}{c|}{\xmark} & \multicolumn{1}{c|}{\xmark} & \multicolumn{1}{c|}{\xmark} & \cmark  \\ \hline
\end{tabular}
\label{table_time}
\end{table}

\subsection{Ablation Study on Kernel Graft}
In this experiment, we evaluate the performance of kernel bypass graft. Specifically, we train our proposed method in two different settings for comparison. One setting is that the model does not adopt any bypass-graft kernel, denoted as w/o Graft. The second is that the model adopts the bypass-graft kernel obtained in the Euclidean space, denoted as E-KMeans. Our proposed method learns the kernel graft in the Wasserstein space, and is denoted as W-KMeans. 
The average PSNR and SSIM for the different settings on the REDS4 dataset are shown in Table \ref{setting}. 
As observed, the model with the graft learned in the Euclidean space achieves comparable results with the model without using grafts, because the grafted kernels learned in the Euclidean space cannot effectively inherit prior information from the extracted kernels. In contrast, our model using the kernel graft learned in the Wasserstein space significantly outperforms other compared settings because the optimal mapping can help the learned kernels better preserve the geometric properties.

\begin{table}[htbp]
\centering
\caption{ The average PSNR and SSIM of our proposed model with different settings on the REDS4 dataset. $\uparrow\bigtriangleup$ denotes the performance increment. The best results are highlighted in bold.}
\begin{tabular}{c|c|c|c}
\hline
 &  w/o Graft & E-KMeans & W-KMeans \\ \hline
PSNR &29.58 & 29.59 & \textbf{29.73} \\ 
$\uparrow\bigtriangleup$ & 0.00 & 0.01 & \textbf{0.15} \\\hline
SSIM & 0.8480 & 0.8484 & \textbf{0.8514} \\ 
$\uparrow\bigtriangleup$ & 0.0000 & 0.0004 & \textbf{0.0034} \\\hline
\end{tabular}
\label{setting}
\end{table}

\subsection{Ablation Study on Different Kernels}
\begin{figure}[htbp]
\centering
    \subfloat[ACB]{
        \includegraphics[width=0.45\linewidth]{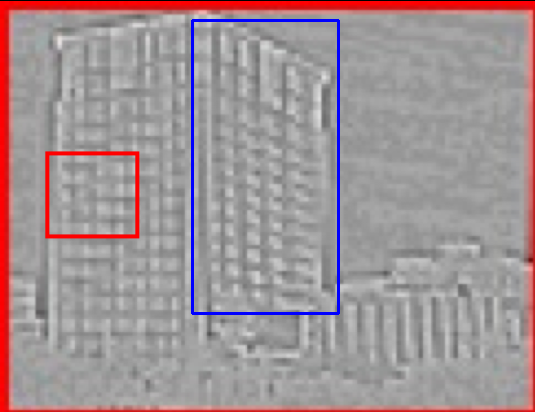}
    }
    \subfloat[RepVGG]{
        \includegraphics[width=0.45\linewidth]{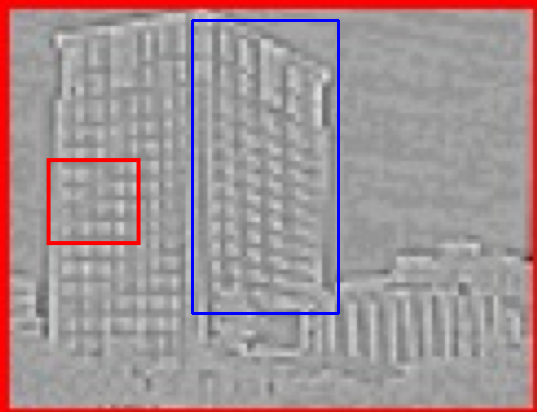}
    } \\
    \subfloat[ECB]{
        \includegraphics[width=0.45\linewidth]{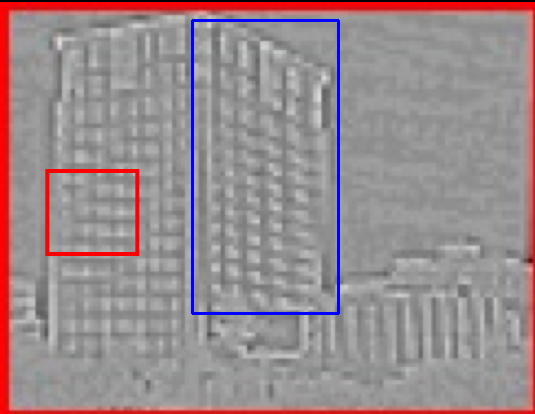}
    }
    \subfloat[Ours]{
        \includegraphics[width=0.45\linewidth]{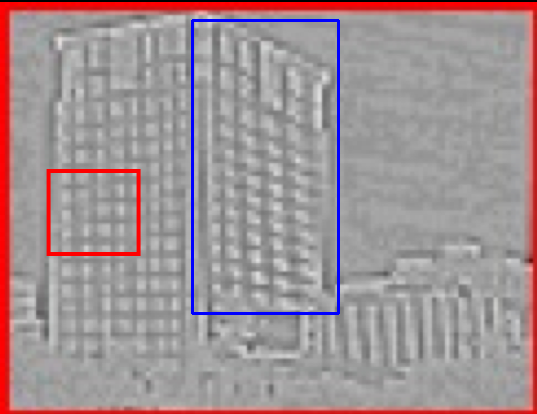}
    }
\caption{The resulting features generated by ACB, RepVGG, ECB, and our proposed method, i.e., CKBG.}
\label{fig_kernels}
\end{figure}

In this experiment, we compared our proposed bypass-grafted kernels with other re-parameterized kernels, such as the ACB kernel \cite{lo2019efficient}, RepVGG kernel \cite{ding2021repvgg}, and ECB kernel \cite{zhang2021edge}. Specifically, we retrained the model by replacing our proposed grafted kernel with other re-parameterized kernels. The average PSNR and SSIM of the different kernel settings on the REDS4 dataset are shown in Table \ref{kernels}. We can find that our method achieves the best performance compared with other kernel settings. Figure \ref{fig_kernels} illustrates the feature maps generated from the last convolutional block of the models with different kernels. We highlighted regions of dense grids with the red and blue rectangular boxes for better visual comparison. As observed, the generated result of our method looks much sharper, and the marked region of dense grids has less distorted content.

\begin{table}[htbp]
\centering
\caption{The average PSNR and SSIM results of different re-parameterization convolutional blocks on the REDS4 dataset. The best results are highlighted in bold.}
\begin{tabular}{c|c|c|c|c}
\hline
 & ACB \cite{lo2019efficient} & RepVGG \cite{ding2021repvgg} & ECB \cite{zhang2021edge} & CKBG(ours) \\ \hline
PSNR & 29.53 & 29.51 & 29.58 & \textbf{29.73} \\ 
SSIM & 0.8471 & 0.8464 & 0.8478 & \textbf{0.8514} \\ \hline
\end{tabular}
\label{kernels}
\end{table}

\subsection{Ablation Study on Knowledge Transfer}
We further compare our method with two deep super-resolution models, i.e, FAKD \cite{he2020fakd} and FDVDNet \cite{xiao2021space}, which are based on knowledge distillation. The compared methods distill prior knowledge from the teacher network based on the feature map generated from the intermediate layers. In addition, the teacher networks provide supervised signals to guide the student networks for training. In contrast, our proposed CKBG method extracts convolutional kernels from a pretrained image super-resolution model and grafts the extracted kernels to a lightweight model before training. The average PSNR and runtime of the different methods on the Vid4 dataset are tabulated in Table \ref{transfer}, where the best results are highlighted in bold. The runtime of the methods is measured in a device equipped with a NVIDIA GeForce GTX 1080 Ti. From Table \ref{transfer}, our method can achieve the best performance with the lowest runtime compared with FAKD and FDVDNet. This shows that our method is more effective and efficient for video super-resolution.
\begin{table}[htbp]
\centering
\caption{The average PSNR and runtime of SR models with different knowledge transfer methods on the Vid4 dataset. The best results are highlighted in bold. The runtime is measured in a Navdia 1080 Ti.}
\begin{tabular}{c|c|c|c|c}
\hline
 & Bicubic \cite{lo2019efficient} & FAKD \cite{he2020fakd} & FDVDNet \cite{xiao2021space} & CKBG(ours) \\ \hline
PSNR & 23.78 & 25.42 & 26.14 & \textbf{26.34} \\ 
Time & - & 28.31\,ms & 17.50\,ms & \textbf{12.26}\,ms \\ \hline
\end{tabular}
\label{transfer}
\end{table}


\section{Conclusion}
Online video applications raise high requirements for video super-resolution (VSR) algorithms in terms of processing latency, model complexity, and distortion quality, but few of the existing VSR methods can handle such challenging issues simultaneously. In this paper, we propose an extremely low-latency VSR method for online applications. A novel knowledge transfer method, called convolutional kernel bypass graft (CKBG), is proposed. The proposed CKBG aims to improve the performance of a base VSR network by bypassing a set of extra kernels containing rich prior knowledge from external, pretrained SR models (i.e., kernel grafts). Experiment results show that our proposed method can process a video sequence at a rate of up to 110 FPS, and achieve the best trade-off between the distortion quality and the processing latency, compared with other competitive VSR methods.


{\appendices
\section{Kernel Re-parameterization of CKBG}
In the paper, we propose a novel kernel knowledge-transfer method, called convolutional kernel graft bypass, which enhances a base VSR model by bypassing the learned ``kernel graft'' from a large pretrained model. The resulting network module is called a bypass graft block (BGB), which has a multi-branch structure for feature extraction. In our method, the proposed BGB is trained in the form of a multi-branch structure, but the multiple branches can be converted into a convolutional operator in the testing stage because of the linear property of the convolutional operation \cite{ding2021repvgg}. This kernel conversion process is called the kernel re-parameterization, as illustrated in Figure \ref{wp-conv}. Our proposed method involves two types of kernel re-parameterization: each sequential convolution and parallel convolution. In the testing stage, the sequence of $1\times 1$ convolution and $3\times 3$ convolution is first merged into a $3\times 3$ convolution. Then, the convolutional kernels of all parallel branches are merged into a single convolution kernel. The resulting structure is a single-path topology. Next, we will elaborate on how to perform kernel re-parameterization for sequential and parallel convolutions.

\textit{Type I: Sequential convolution.} Suppose a sequential convolution of a $1 \times 1$ kernel and a $K\times K$ kernel, denoted as $F_{1}\in\mathbb{R}^{\text{C}_{in}\times \text{C}_{mid}\times 1\times 1}$ and $F_{2}\in\mathbb{R}^{\text{C}_{mid}\times \text{C}_{out}\times K\times K}$, respectively, where $\text{C}_{in}$, $\text{C}_{mid}$, and $\text{C}_{out}$ represent the number of channel of the input feature, intermediate feature, and output feature, respectively. The output of the sequential convolution is computed as follows:
\begin{align}
    F_{output} &= (F_{in}\ast F_{1} + \mathbf{b}_{1})\ast F_{2}+ \mathbf{b}_{2} \\
    &=F_{in}\ast F_{1}\ast F_{2} + \mathbf{b}_{1}\ast F_{2} + \mathbf{b}_{2}, \label{seq-conv}
\end{align}
where $\ast$ denotes the convolutional operation, and  $\mathbf{b}_{1}\in\mathbb{R}^{1\times C_{mid}}$ and $\mathbf{b}_{2}\in\mathbb{R}^{1\times C_{out}}$ are the corresponding bias terms. According to the linear property of convolution, the convolution kernels $F_{1}$ and $F_{2}$ in the first term can be merged as follows:
\begin{equation}
    F^{\prime} = F_{2}\ast F^{T}_{1},
\end{equation}
where $F^{T}_{1}\in \mathbb{R}^{C_{mid}\times C_{in}\times 1\times 1}$ is the transpose of $F_{1}$, and $F^{\prime}$ denotes the resulting convolution kernel with the size of $\text{C}_{in}\times \text{C}_{out}\times K\times K$. For the second term $\mathbf{b}_{1}\ast F_{2}$ , the corresponding re-parameterized result  $\hat{\mathbf{b}}_{1}=[\hat{b}_{1,1},\cdots, \hat{b}_{1, C_{mid}}]$ is computed as follows:
\begin{equation}
    \hat{b}_{1,c_{out}} = \sum_{c =1}^{C_{mid}}\sum_{m=1}^{K}\sum_{n=1}^{K}b_{1,c}F_{c_{out}, c, m, n}^{(2)},
\end{equation}
where $c_{out}=1,\cdots, C_{out}$, and $F_{c_{out}, c, m, n}^{2}$ represents the kernel at the position $(c_{out}, c, m, n)$ of the convolution kernel $F_{2}$. Then, the final re-parameterized bias term is obtained as follows
\begin{equation}
    \hat{\mathbf{b}}=\hat{\mathbf{b}}_{1} + \mathbf{b}_{2}. 
\end{equation}
As a result, the sequential convolution of a $1\times 1$ convolution and a $K\times K$ convolution is merged into a single convolutional kernel after performing kernel re-parameterization. Eqn.\,\ref{seq-conv} can be re-written as follows:
\begin{equation}
    F_{out} = F_{in} \ast F^{\prime} +\hat{\mathbf{b}}.
\end{equation}
\begin{figure*}[htbp]
    \centering
    \includegraphics[width=0.9\linewidth]{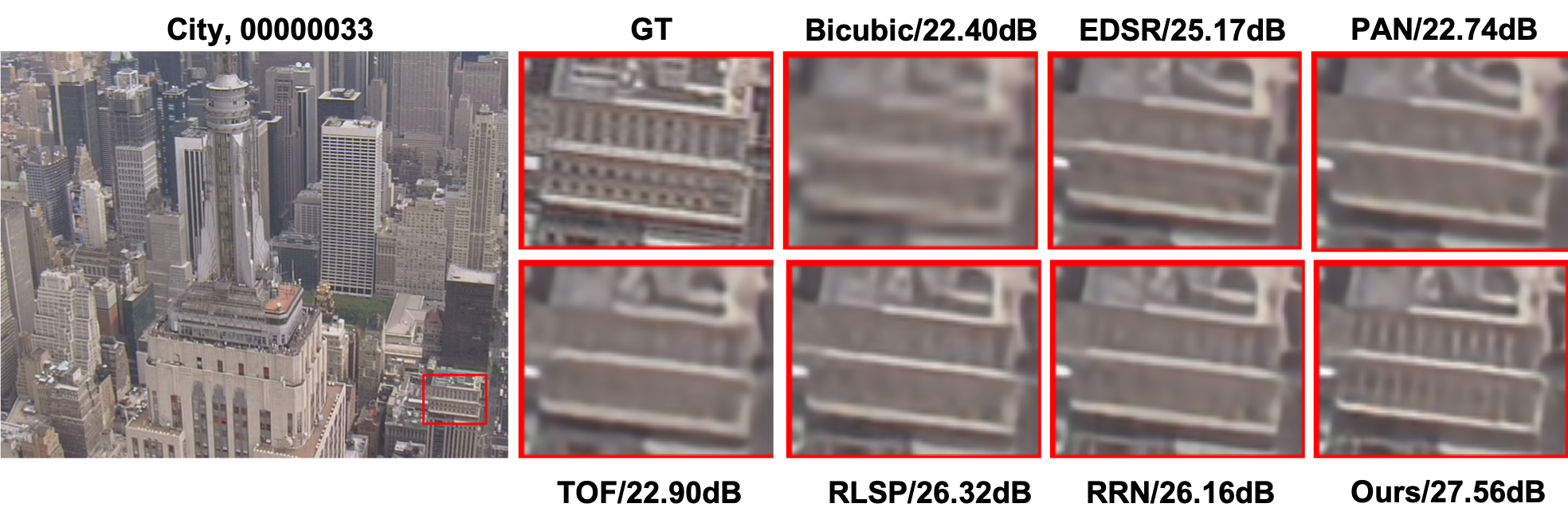}
    \caption{The illustrated image ``City 00000033" is selected from the Vid4 dataset. The region marked by the red box is generated by different VSR methods for visual comparison. }
    \label{visual_results3}
\end{figure*}
\begin{figure*}[htbp]
    \centering
    \includegraphics[width=0.9\linewidth]{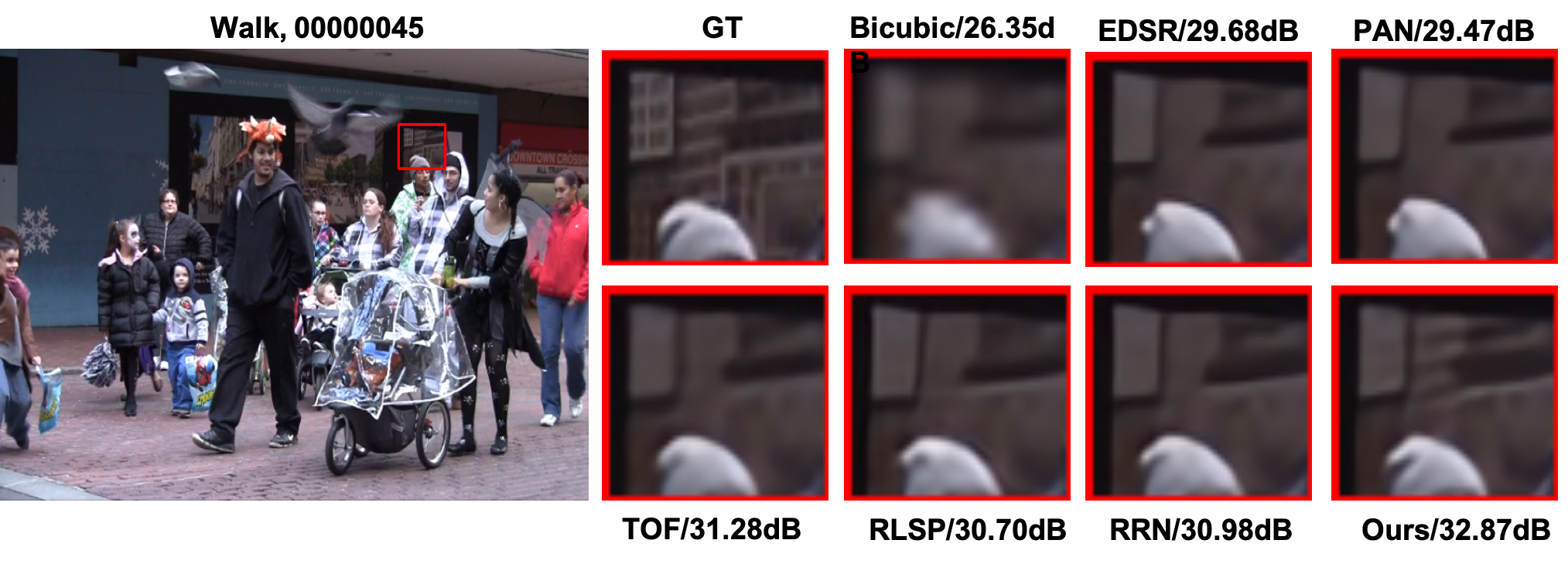}
    \caption{The illustrated image “Walk 00000045” is selected from the Vid4 dataset. The region marked by the red box is generated by different VSR methods for visual comparison.}
    \label{visual_results4}
\end{figure*}

\textit{Type II: Parallel convolution.} Assume that the parameter set of the $i$-th branch in the $\ell$-th BGB is denoted as $\Theta^{\ell}_{i}=\{F^{\ell}_{i}, \mathbf{b}^{\ell}_{i}\}$, where $F^{\ell}_{i}\in\mathbb{R}^{C_{in}\times C_{out}\times K\times K}$ and $\mathbf{b}\in\mathbb{R}^{1\times C_{out}}$ represent the convolution kernel and the bias term, respectively. Without loss of generality, suppose a BGB has $B$ branches. Given an input feature map $F_{in}^{\ell}$, the output of the multi-branch block is computed as follows:
\begin{equation}
    F_{out}^{\ell} = \sum_{i=1}^{B}(F_{in}^{\ell}\ast F_{i}^{\ell} + \mathbf{b}_{i}^{\ell}).\label{conv-eq} 
\end{equation}
Then, we can perform kernel re-parameterization based on the linear property of convolution again. As a result, Eqn.\,(\ref{conv-eq}) can be re-written as follows:
\begin{align}
    F_{out}^{\ell} &= F_{in}^{\ell}\ast \left(\sum_{b=1}^{B} F_{b}^{\ell}\right) +\left(\sum_{b=1}^{B} \mathbf{b}_{b}^{\ell})\right) \\
    &= F_{in}^{\ell} \hat{F}^{\ell} + \hat{\mathbf{b}}^{\ell},
\end{align}
where $\hat{F}^{\ell}=\sum_{b=1}^{B}F_{b}^{\ell}$ and $\hat{\mathbf{b}}^{\ell}=\sum_{b=1}^{B}\mathbf{b}_{j}^{\ell}$ are the corresponding re-parameterized convolutional kernel and the bias term. As a result, the multi-branch structure is converted into a single-path structure in the testing stage. 

Since kernel re-parameterization relies on the linear property of the convolution operation, the re-parameterized kernel in the testing stage is equivalent to the multi-branch structure in the training stage, without sacrificing any performance.

\section{Implementation Details of BasicVSR$^{\ast}$}
BasicVSR \cite{chan2021basicvsr} is a bi-directional model and has shown its effectiveness for VSR. However, BasicVSR cannot be applied to online video applications because it requires access to the whole video sequence, which is impractical. To make BasicVSR meet the online requirements, we modify BasicVSR in this paper, denoted as BasicVSR$^{\ast}$. Specifically, we first remove the backward part of the model, so it does not need to cache the future frames. To meet the requirement of low model complexity, we reduce the number of residual blocks from 60 to 15. The number of feature channels is reduced from 64 to 32. Other configurations are remained the same as the original model for training.

\section{More Visual Results}
In Figure \ref{visual_results3} and \ref{visual_results4}, we provide additional visual results generated by different image/video SR methods for visual comparison. As illustrated, our proposed method has a better ability to preserve the shapes and textures of objects in the images than other compared methods. In addition, the images generated by our proposed method contain less distorted content than that produced by other compared methods, leading to the best visual quality.
}

 
%

\bibliographystyle{IEEEtran}
\bibliography{refs.bib}


 





\end{document}